\documentclass{article}
\usepackage{spconf,amsmath,graphicx}

\usepackage[ruled,vlined]{algorithm2e}
\usepackage{xcolor}
\usepackage{tabularx}
\usepackage{amsfonts}
\usepackage{enumitem}

\DeclareMathOperator*{\E}{\mathbb{E}}

\title{Sensor Adversarial Traits: Analyzing Robustness of 3D Object Detection Sensor Fusion Models}
\name{Won Park$^{\star}$ \qquad Nan Liu$^{\star}$ \qquad Qi Alfred Chen$^{\dagger}$ 
\qquad Z. Morley Mao$^{\star}$}
  
 \address{$^{\star}$ University of Michigan \\
       $^{\dagger}$ UC Irvine}

\begin{document}
%
\maketitle

\addtolength{\topskip}{-1cm}
\addtolength{\parindent}{-1mm}
\setlength{\parsep}{0pt}
\setlength{\labelwidth}{8pt}
\setlength{\itemsep}{0pt}
\addtolength{\parskip}{-0.2mm}
\addtolength{\abovedisplayskip}{-0.1cm}
\addtolength{\belowdisplayskip}{-0.1cm}

\begin{abstract}
A critical aspect of autonomous vehicles (AVs) is the object detection stage, which is increasingly being performed with \textit{sensor  fusion  models}: multimodal 3D object detection models which utilize both 2D RGB image data and 3D data from a LIDAR sensor as inputs.   
In this work, we perform the first study to analyze the robustness of a high-performance, open source sensor fusion model architecture towards adversarial attacks and challenge the popular belief that the use of additional sensors automatically mitigate the risk of adversarial attacks.
We find that despite the use of a LIDAR sensor, the model is vulnerable to our purposefully crafted image-based adversarial attacks including disappearance, universal patch, and spoofing. After identifying the underlying reason, we explore some potential defenses and provide some recommendations for improved sensor fusion models. 
  

\end{abstract}
\section{\centering Introduction}

Autonomous vehicle (AV) manufacturers  often use  \textit{sensor fusion} models to help vehicles detect the environment around them.
These types of models are multimodal 3D object detection models that take in two types of inputs: a 2D image from a camera and 3D depth data usually from a LIDAR sensor. 
With the growing proliferation of autonomous vehicles, their security is becoming more paramount, especially against adversarial examples ~\cite{CCS_Physical, sun2020lidar}.  

It has long been known in the community that machine learning models are vulnerable to adversarial examples, maliciously crafted inputs designed to intentionally fool the model into outputting an erroneous result. 
These range from attacks in the raw pixel space \cite{Xie_2017} to launching these attacks in the physical world ~\cite{EykholtWOOT, CCS_Physical, huang2019upc, personpatch}. 

There is a belief that the use of additional inputs can mitigate the effect of adversarial examples. 
While recent work~\cite{Kim2019} has shown theoretically that models that take in multiple inputs are  vulnerable to potential perturbations in a single input, no one has actively explored the robustness and crafted adversarial examples against sensor fusion models.
Our work is the first to demonstrate the insecurity of sensor fusion models to several  realistic adversarial attacks for 3D object detection. 

Though there are many multimodal 3D detection architectures available, we focus this study on models that take in the two types of inputs simultaneously.
We purposefully choose to ignore model architectures such as Frustum-Pointnet ~\cite{Fpointmnet} that utilize a "pipeline" structure in which image data is taken in first followed by LIDAR data because these models are trivially vulnerable to image-based attacks --- any existing attack algorithm to fool 2D image object detectors will be able to fool the entire model. 
Instead, for this work, we choose AVOD ~\cite{AVOD}, an open-source 3D object detection model, because of its near-top performances among open-source models in the KITTI benchmark.
Note that this model differs from a prior work~\cite{wang2020robust} that does not evaluate attacks on a properly created sensor fusion model whose architecture is conducive to 3D multimodal object detection.
Instead, they simply combine existing architectures - a LIDAR featurizer with a YOLO model, for example.
The difference between the architectures is made apparent in the AP scores reported: AVOD, studied here, reports 71.88 while the paper's architecture has a score of 60.3.
In short, our attacks are more meaningful because we evaluate and attack a model with a higher baseline accuracy.
Secondly, whereas the aforementioned work only explores one attack, we are able to develop a wide variety of attacks and delve deeper into the nature of sensor fusion models. 



We are interested in understanding if the use of an additional input prevents adversarial attacks on the other input. 
Though we could utilize adversarial attacks on the LIDAR input like previous work~\cite{ccs:yulong, sun2020lidar, DBLP:journals/corr/abs-1912-11171, Tu_2020_CVPR}, we instead choose to focus on modifying the image input. 
This is because we find that the model relies more heavily on LIDAR data and successful attacks using modification of just the LIDAR is more trivial.
Furthermore, physical attacks on the camera detector are  more realistic and potent than the ones on the LIDAR sensor. 
Thus, by restricting our attacks to just images, we are assuming a less powerful and more realistic attacker.
Because this is the first foray into this field, we assume that the adversary is a white-box attacker, having full access to the model. 
Despite this, in order to guide future research works, we aim to be as realistic as possible; 
this includes restrictions that the adversary will not be able to modify the model arbitrarily, including any post-processing steps.

Our key contributions are as follows:
\setlist[itemize]{leftmargin=*}
\begin{itemize}
\itemsep=-.5em 
\item We perform the first study of adversarial examples on proper sensor fusion models for 3D object detection.
We modify existing techniques to show that sensor fusion models are vulnerable to adversarial attacks that modify just the image input. 
These attacks include the \textit{raw pixel disappearance attack} (94.17\% success rate) and a spoofing attack (89.1\%). 
We then analyze the model architecture to show that despite the symmetric architecture, the model frequently leans heavily on the LIDAR input to detect obstacles. 

\item Building upon the raw pixel disappearance attack, we develop a new methodology of constructing generalized adversarial examples in which one single noise can fool many samples. 

\item We explore some basic defenses, including robust training and a novel fusion layer ~\cite{Kim2019}. We comment on their effectiveness and put forth suggestions for future directions.

\end{itemize}


\section{\centering Crafting Attacks}
\label{section:disappear}
\begin{figure}
\begin{center}
\includegraphics[scale = 0.42]{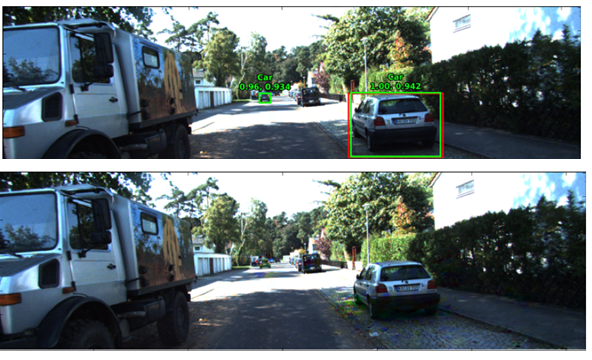}
\end{center}
  \caption{Results of some of our disappearance attacks. Top is benign images and bottom is adversarial images. The 1st value corresponds to the classification confidence and the 2nd value corresponds to the IOU with the ground truth bounding box. The red boxes are ground truth and the green boxes are bounding boxes outputted by the model. }
\label{fig:succesful_adv_attack}
\end{figure}

In this section, we explore attacks that are able to fool the model into not detecting an object it had previously detected (\textit{raw-pixel disappearance attacks}) and those that fool the model into detecting an object that is not actually present (\textit{spoofing attack}).


The \textit{raw-pixel disappearance attack} is motivated by a desire to create a disappearance attack that results in an adversarial example that is less noticeable to the human eye.
We explore a different kind of attack - patch attacks - in Section \ref{section:physical}. 
To cause a desired object to disappear, we aim to force the output softmax probabilities of all potential bounding boxes around the said object below the detection threshold.
We will call this set of all potential boxes that we need to attack $B$. 
For ease of notation, suppose $C(w, b) \in R^c$ denotes the output classification softmax of bounding box $b$ on image $w$ and $C(w)$ outputs all the potential bounding boxes of image $w$ in decreasing order according to softmax score.
We attempt to find an adversarial sample $\delta$ that minimizes the following function:
\begin{equation}
L(w+\delta, B) =   \sum_{b \in B} [C(w+\delta, b) ] + \epsilon * D(w+\delta, w ) 
\end{equation}

 The second element $D()$, which measures the $L_2$ norm between the adversarial image and the regular image,  is added to make the perturbation to the image as small as possible, as suggested by the CW attack ~\cite{Carlini_2017}.
The optimal value of the weighting coefficient $\epsilon$ is found through binary search.
Unlike previous work, we choose to attack the second-stage detector (instead of the first stage RPN) as it results in an attack with less distortion. 

However, due to NMS and the restrictions we set on the adversary, not all of the elements of $B$ will be visible and we are unable to know  the set $B$ a priori.
For some related previous work, this was not a huge limiting factor ~\cite{BMVC} while others went around this by trying to attack the NMS algorithm itself or to modify the NMS threshold to obtain all bounding boxes ~\cite{Xie_2017}.

\begin{figure}
\centering
\includegraphics[scale = 0.27]{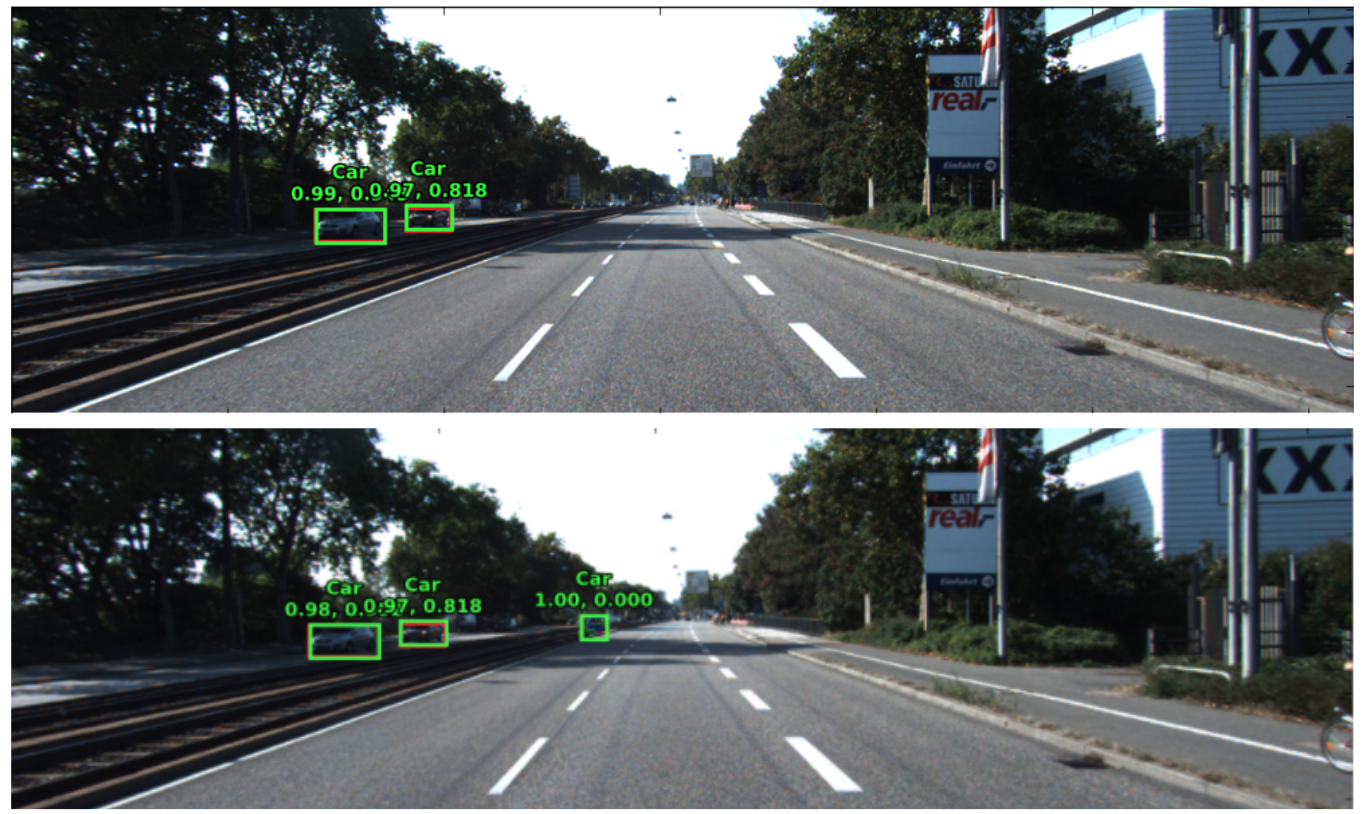}
  \caption{Results of some of our spoofing attacks. Top is benign and bottom is adversarial. bounding boxes with an IOU of 0 (second value) are the spoofed objects.}
\label{fig:spoof_attack}
\end{figure}




Since neither solution is allowed under our threat model, we instead modify the algorithm to greedily attack the top confidence bounding box that is visible - as we keep trying to lower the confidence of the bounding box with the highest score, one of two outcomes will happen. 
In one, the object in question will no longer be detected, in which case our attack goal is accomplished. 
In the other case, the bounding box in question will be removed via NMS and the next top-score bounding box will appear and the process can be repeated.
This process will remove \text{all} objects present in an image, but an adversary can selectively remove certain objects by applying a mask.
In this case, the objective function needs to be modified to attack the top $k$ bounding boxes simultaneously.

For the spoofing attack, in addition to switching the sign of the loss function, we add another element to our adversarial objective function that will help ensure the desired bounding box is outputted by the RPN.
The final loss function is a weighting of the two losses where the RPN loss function is weighted much more heavily:

\addtolength{\abovedisplayskip}{-4cm}
\addtolength{\belowdisplayskip}{-0.1cm}

\begin{equation}
L_{spoof} = L_{RPN} + \alpha * L_{Stage2}
\end{equation}

It is important to note that a simple defense against spoofing attacks is to remove any anchor boxes that do not have any LIDAR data as a pre-processing step before inputting into the model.
Acknowledging this, we run all of our experiments under this assumption to increase the likelihood that our results cannot be defeated by this simple defense.

\subsection{Evaluation and Results}

When training AVOD, we follow the methodology followed in the paper proposed by Chen et al ~\cite{MV3D} on the KITTI dataset ~\cite{Geiger2013IJRR},
We split the trainval set into a training set with 3712 samples and a validation set with 3769 samples.
We train all models to closely match the results stated in the original paper.

For our experiment, we choose 3000 random samples containing a total of 10,920 detected vehicles and 7,585 pedestrians / cyclists and try constructing adversarial examples using the method stated above.
We are able to achieve a 94.17\% success rate on vehicles and 97.11\% on pedestrians and cyclists.
Some of the attacks are shown in Figure \ref{fig:succesful_adv_attack}. 
For the spoofing attack, we are able to achieve a 89.1\%  and 91.33\%, respectively, upon evaluating on these samples \ref{fig:spoof_attack}.


\section{\centering Towards Generalizability}
\label{section:physical}

In this section, we attempt to create a single adversarial patch that, despite being more noticeable to the human eye, would be able to be universally applied to any vehicleand cause them to escape detection from the model.
This is a key step in determining the feasibility of attacks in the physical world. 

To start, we draw inspiration from the expectation over transformation (EOT) algorithm \cite{athalye2017synthesizing}. 
Due to the difficulty of applying any transformation to an image and also properly modify the corresponding LIDAR data, however, we use different object samples available in KITTI instead as input. 
For each image, we identify an area on the vehicle we wish to apply a patch. 
Note that if an image has multiple objects, we may have to apply the patch separately to different areas.
Let $P(w,\delta)$ be the operation that applies adversarial patch $\delta$ to image $w$, appropriately resizing the patch as necessary.
If we have a set of images $T$ (along with their corresponding bounding box set), we would be able to create a universal patch by solving the following objective function:

\begin{equation}
\text{argmin} _\delta  \E _{w \in T}[ L( P(w,\delta)) ]
\end{equation}

Normally, this would be done simultaneously via batching.
However, since AVOD and many other sensor fusion models do not support batching, we alter the algorithm:
instead of operating over all the images simultaneously, we perform the objective function on one image at a time, keeping the noise in between images and iterate over all the images trying to ensure convergence.
The number of times to iterate is a hyperparameter that must be tuned but for our experiments we iterated 25 times. 
This is a similar approach as suggested by \cite{UniversalAP}, however, we do not project all perturbations onto a p-norm since we find it slows our algorithm due to the nature of our loss function trying to target every potential bounding box.
The resizing and the nature of multiple bounding boxes are also reasons why the GD-UAP\cite{gd-uap} is less than ideal for this work. 
Therefore, we take advantage of updating $\epsilon$ to control the distortion.
We find that gradually decreasing the value until a certain floor value works well.
The algorithm can be viewed in \ref{alg:modifiedEOT}.

\begin{algorithm} 
\SetAlgoLined
\SetKwInOut{Input}{input}\SetKwInOut{Output}{output}
\Input{Set of images $T$, k, n, $\epsilon$}
\Output{Adversarial noise $\delta$}
\Begin{
$\delta \longleftarrow \text{RandomInit} $\;
 \For{$i = 0$  \KwTo $n$}{
  $w \leftarrow \text{NextImage}(T, i) $\;
  $B' \leftarrow C(P(w,\delta)) [0...k]$\;
  $ \delta \leftarrow argmin _\delta L(P(w,\delta), B', \epsilon) $
  $ \epsilon \leftarrow \text{UpdateEpsilon}(i)$
  
 }
 \Return $\delta$
}
 \caption{Modified EOT}
 \label{alg:modifiedEOT}

 \end{algorithm}

\subsection{Results}
We run this algorithm on the validation set and are able to achieve a success rate of 64.03\%.
To establish a baseline comparison, we apply a random noise patch to the same vehicles.
These random noise patches, when applied to the vehicle in the same location, achieve a 0\% success rate.
The results of this case study suggests a worrisome fact that sensor fusion models are still vulnerable to universal physical adversarial examples, similar to what is shown in Huang et al \cite{huang2019upc}.

\section{\centering Analysis of Sensor Input}
\label{sec:input}

Motivated by the results of our experiments on various attacks, we suspect that the model architecture, while symmetrical, heavily utilizes the LIDAR sensor input over the image.
To test this, we run an experiment in which we use the LIDAR from one sample and the image for another to understand how the model performs when the image and the LIDAR are at odds with each other. 
This was done for 600 random samples, swapping the image of one and the LIDAR of another, resulting in 360,000 combinations. 
For the sake of simplicity, we consider an object as "correctly identified" if the bounding box was correctly drawn according to the LIDAR sensor
~\footnote{Note that we could have easily swapped and used the image input as "ground truth" but that would not make any change to the final result}. 
Amongst all the potential bounding boxes, 91\% were correctly identified, despite having conflicting image data. 
Furthermore, only 19\% of all bounding boxes detected did not correspond to any ground truth bounding box.  

This experiment strongly suggests that the model favors LIDAR data when detecting objects, which helps explain the difficulty in the spoofing attacks. 
For instance, when we compare the L2 norm per pixel, we find that the spoofing attack requires much more distortion: while the disappearance attack required a median per-pixel distortion of roughly 0.28 in L2 space, the spoofing attack required a L2 distortion of over 1.3 in L2 space. 
This is line with a previous work that found another sensor fusion model, MV3D, also favors LIDAR ~\cite{sun2020lidar}.
While this is contrary to what was found in Wang et al ~\cite{wang2020robust}, we believe this is because their lack of a true sensor fusion model built from the ground up. 
 This also suggests that the use of image in this architecture proves to be an "Achilles' heel": while most of the detection of an object is done using the LIDAR input, it is not sufficient, as the image provides a way for adversaries to override this and attack the model. 

\section{\centering Exploring Defenses}
\label{section:defense}

\begin{table}
\centering
\begin{tabular}{ |c|c|c| } 
 \hline
 Type & Disappearance  & Spoof \\
 \hline\hline
 Baseline & 0.94 & 0.89 \\
 Distorted Inputs & 0.92 & 0.85\\ 
 MaxSSN  \cite{Kim2019}&  0.87 & 0.81\\ 
  MaxSSN + LEL \cite{Kim2019} &  0.80   &  0.39\\ 
 Adversarial Training &  0.63   &  0.51\\ 
 \hline

\end{tabular}
\caption{Table showing the success rate (in \%) of our attacks against various defenses.}
\label{tab:defense_results}
\end{table}

In this section we analyze some potential defenses against adversarial examples on sensor fusion models. 
For each defense, we test our raw-pixel disappearance attack and our spoofing attack on the same vehicle samples as in the evaluation for our attacks. 
All results can be seen in Table \ref{tab:defense_results}.

We first attempt to address a belief that training on a wider range of inputs (like fog and snow) will help mitigate adversarial examples.
We train an instance of AVOD on an augmented version of the dataset in which we apply all distortions as suggested by Hendryks and Dietterich \cite{hendrycks2018} to every training image.
While training on various input distortions may be important for safe performance of AVs, we find that it does not eliminate the threat against adversarial examples. 

We next analyze methodologies proposed by Kim and Ghosh \cite{Kim2019} that provide robustness against single-source distortion in sensor fusion models.
We refer readers to the paper for details, but in short, the authors propose novel loss functions and a new fusion layer called LEL to protect against noisy distortion. 
We test the two designs proposed that achieved the best performance under noise: using the new loss function called MaxSSN with and without LEL.
We train both models according to the specifications shown in the paper and achieve a similarly stated accuracy.
We find that the models do mitigate against our disappearance attacks, but only to a success rate of around 80\%.
However, the addition of the LEL does better in defending against spoofing attacks.
Unfortunately, it is worth noting that there exists a trade-off as both models suffer in a drop of AP score compared to the original model when run on benign inputs. 

A popular methodology to protect against adversarial examples is adversarial training. 
We utilize a preliminary technique proposed by Zhang and Wang \cite{adv_training}.
We find that the AP score  drops  after adversarial training, but so does the success rate of the adversarial attacks drop as well; the raw-pixel disappearance attack drops to a 63\% effectiveness while the spoofing attack drops to a 51\% success rate.
While these results are not ideal, they do not necessarily eliminate adversarial training as a viable option to provide robustness. 
On the contrary, they demonstrate that a better adversarial training algorithm may be able to provide robustness. 
We leave this exploration to future work.
We also believe another avenue to explore is incorporating defenses \textit{outside} the model. 
In a larger system, it could be feasible to utilize the different sensors, for example, to validate one another before feeding into the final detection model.
We leave these possible defenses for future work to explore.

\section{\centering Conclusion}
In this paper we explore a fusion model's security against adversarial examples.
We discover that the use of a secondary input provides limited defense against a myriad of different adversarial attacks. 
Though we only evaluate on one model due to the lack of availability of open-source models, our evaluation on alternative fusion layers and training loss functions suggest that other models may also be vulnerable to single image attacks.
We urge future works on sensor fusion models to help increase robustness. 
\section{\centering Acknowledgements}
We thank our reviewers for their comments. We also wish to thank Fahad Kamran, Jiachen Sun, Yulong Cao for their help and insights.
This project is partially supported by Mcity and NSF under the grants CMMI-2038215, CNS-1930041, CCF-1628991, CNS-1544678, CNS-1850533, CNS-1929771, and CNS-1932464.

\bibliographystyle{IEEEbib}
\bibliography{egbib}

\end{document}